\documentclass[conference]{IEEEtran}
\IEEEoverridecommandlockouts

\usepackage{cite}
\usepackage{amsmath,amssymb,amsfonts}
\usepackage{algorithmic}
\usepackage{graphicx}
\usepackage{textcomp}

\def\BibTeX{{\rm B\kern-.05em{\sc i\kern-.025em b}\kern-.08em
    T\kern-.1667em\lower.7ex\hbox{E}\kern-.125emX}}

\usepackage[utf8]{inputenc} 
\usepackage[T1]{fontenc}    
\usepackage{graphicx}       

\usepackage{multirow}       
\usepackage{kotex}          
\usepackage{xcolor}         
\usepackage{amsmath}        
\usepackage{amssymb}        
\usepackage{microtype}      
\usepackage{booktabs}       
\usepackage{makecell}       
\usepackage{colortbl}  
\usepackage[colorlinks=true, linkcolor=red, citecolor=blue!60, urlcolor=magenta]{hyperref} 

\definecolor{lightblue}{HTML}{E8F0FE}

\newcommand{\rc}[0]{\rowcolor{lightblue}}

\newcommand{\sref}[1]{Section \ref{#1}}
\newcommand{\fref}[1]{Fig.~\ref{#1}}
\newcommand{\eref}[1]{Eq.~(\ref{#1})}
\newcommand{\tref}[1]{Table~\ref{#1}}

\title{D-Cube: Exploiting Hyper-Features of Diffusion Model for Robust Medical Classification}

\author{
    \IEEEauthorblockN{Minhee Jang\IEEEauthorrefmark{1}\thanks{\IEEEauthorrefmark{1}Minhee Jang and Juheon Son are co-first authors.}, Juheon Son\IEEEauthorrefmark{1}, Thanaporn Viriyasaranon\IEEEauthorrefmark{3}, Junho Kim\IEEEauthorrefmark{4}\IEEEauthorrefmark{2}\thanks{\IEEEauthorrefmark{2}Corresponding author: Junho Kim, Jang-Hwan Choi}, Jang-Hwan Choi\IEEEauthorrefmark{3}\IEEEauthorrefmark{2}}
    \IEEEauthorblockA{\IEEEauthorrefmark{1}Artificial Intelligence Convergence, Departments of Artificial Intelligence and Software, Ewha Womans University, Seoul, Korea}    \IEEEauthorblockA{\IEEEauthorrefmark{3}Department of Artificial Intelligence, Ewha Womans University, Seoul, Korea}
    \IEEEauthorblockA{\IEEEauthorrefmark{4}NAVER AI Lab}
    \{minhee\_jang, thswngjs77, choij\}@ewha.ac.kr, thanaporn.v09@gmail.com, jhkim.ai@navercorp.com
}

\begin{document}

\maketitle

\begin{abstract}
The integration of deep learning technologies in medical imaging aims to enhance the efficiency and accuracy of cancer diagnosis, particularly for pancreatic and breast cancers, which present significant diagnostic challenges due to their high mortality rates and complex imaging characteristics. This paper introduces \textit{Diffusion-Driven Diagnosis (D-Cube)}, a novel approach that leverages hyper-features from a diffusion model combined with contrastive learning to improve cancer diagnosis. D-Cube employs advanced \textit{feature selection} techniques that utilize the robust representational capabilities of diffusion models, enhancing classification performance on medical datasets under challenging conditions such as data imbalance and limited sample availability. The feature selection process optimizes the extraction of clinically relevant features, significantly improving classification accuracy and demonstrating resilience in imbalanced and limited data scenarios. Experimental results validate the effectiveness of D-Cube across multiple medical imaging modalities, including CT, MRI, and X-ray, showing superior performance compared to existing baseline models. D-Cube represents a new strategy in cancer detection, employing advanced deep learning techniques to achieve state-of-the-art diagnostic accuracy and efficiency. The code is available at the provided link\footnote{\url{https://github.com/medical-ai-cv/D-Cube.git}}.
\end{abstract}


\section{Introduction}
Deep learning has become a crucial component in computer-aided decision support systems (CADs) for medical image analysis \cite{Litjens, Shin}. It has significantly improved the efficiency of medical diagnoses, supported treatment processes, and mitigated diagnostic discrepancies among radiologists and physicians. However, despite these advancements, deep learning for medical image diagnosis still faces several challenges.

The development of deep learning for medical image diagnosis faces several obstacles. One major challenge is the need for large annotated datasets to train and validate deep learning models, which is often limited by the time-consuming and costly nature of the annotation process. Additionally, medical image datasets frequently exhibit class imbalances, introducing biases into the models. Each imaging modality, such as computed tomography (CT), magnetic resonance imaging (MRI) \cite{chen2020synthetic,mri2019overview}, and X-ray, presents unique characteristics and complexities, further complicating model development.

\begin{table}[t]
\centering
\caption{\textbf{Performance comparison} based on adaptation Methods in foundation Models on the Pancreas cancer CT Dataset.}
\label{tab:motivation}
\begin{tabular}{l cccc}
\toprule
\multirow{1}{*}{\textbf{Model}}
& \textbf{Acc} & \textbf{F1} \\
\midrule
RADIO (full-tuning)  & 91.00 & 85.10 \\
RADIO + LoRA (fine-tuning)  & 87.20 & 78.50 \\
RADIO + Linear probing & 76.30 & 58.10 \\
Med-CLIP + Linear probing  & 77.96  & 63.59 \\
\midrule
\rc \textbf{D-Cube} &  \textbf{93.61} &  \textbf{89.69} \\

\bottomrule
\end{tabular}
\end{table}

Standard approaches to address these challenges include data sampling techniques like undersampling to balance class representation. However, these methods have shown limited success in improving accuracy or enhancing feature extraction capabilities \cite{Wongvorachan}. Another approach involves using models pretrained on large-scale natural image datasets like ImageNet \cite{BIT, Yamada}. Recently, multimodal foundation models such as RADIO \cite{Ranzinger}, CLIP \cite{Radford}, and DINO \cite{Dino} have shown promising results on downstream tasks using techniques like linear probing or Low-Rank Adaptation (LoRA) \cite{Hu2022LoRA}.

Despite these advancements, models pretrained on natural images often perform suboptimally when applied to medical imaging tasks, as illustrated in Table \ref{tab:motivation}. Our initial studies fine-tuning the multimodal foundation model RADIO on a pancreatic cancer CT dataset showed that these models could not outperform those pretrained specifically on medical images. This performance gap highlights the importance of tailoring models to the specific characteristics of medical image modalities.

To further investigate, we evaluated the Med-CLIP model \cite{medclip}, pretrained on X-ray images, for a CT pancreatic cancer classification task. The results demonstrated suboptimal performance, confirming that image modality significantly impacts model effectiveness. This underscores the necessity of developing models pretrained on medical images tailored to specific modalities to enhance classification performance.

Improving feature extraction quality is critical in addressing medical image analysis challenges. Diffusion models have recently shown promising performance in generative tasks due to their robust mathematical foundations, which prevent convergence to local minima and allow continuous evolution by systematically adding and removing noise. This process helps capture complex structures and inherent data variability, leading to high-quality feature representations. Models like Sora \cite{openai2024sora} and DALL-E \cite{dalle} have demonstrated significant advancements in this area.

This study introduces D-Cube, a novel approach combining contrastive learning with diffusion models. By leveraging high-quality feature representations from diffusion models pretrained on specific medical image modalities, D-Cube aims to address data imbalance and indistinct class boundaries. Our feature selection methodology isolates the most discriminative features, significantly improving classification performance. D-Cube not only performs consistently well across various modalities but also enhances predictive accuracy in medical applications.

We evaluated D-Cube's performance on multiple medical image modalities, including CT pancreatic cancer classification, MRI breast cancer classification, and X-ray COVID-19 classification. The results show that D-Cube outperforms state-of-the-art models across all image modalities.

Our contributions can be summarized as follows:

\begin{itemize}
  \item We introduce a \textbf{novel feature selection technique based on the Gaussianity metric} for diffusion models, leading to significant performance improvements across various medical imaging datasets.
  \item  Our method demonstrates \textbf{robust performance in scenarios with imbalanced and limited datasets}, showcasing strong adaptability and achieving state-of-the-art performance compared to 12 baseline models.
  \item We validate our approach across \textbf{multiple medical imaging domains}, including CT from an Asian population, MRI from a Western population, and X-ray, ensuring its broad applicability.
  \item We present a \textbf{comprehensive ablation study} that highlights the impact of our feature selection method, the use of sub-features from pretrained models, and the integration of tailored loss functions, illustrating the key components contributing to performance gains.
\end{itemize}
\begin{figure*}[t]
    \centerline{\includegraphics[width=0.9\textwidth]{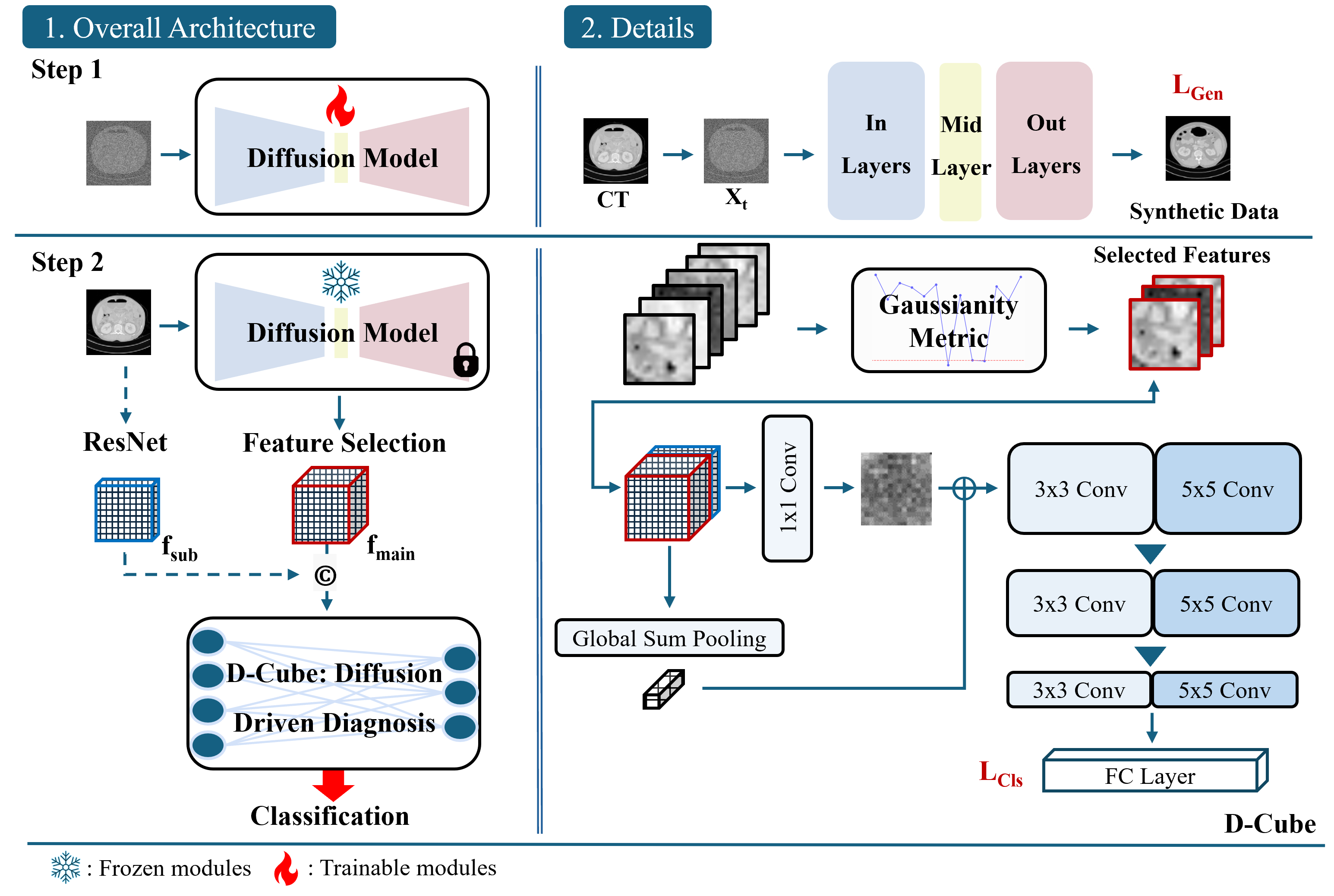}}
    \caption{
    \textbf{Overall Architecture:} Step 1 involves the process of generating diffusion features that enhance the performance of D-Cube, while step 2 entails training a D-Cube for cancer diagnosis using the features generated from frozen diffusion model. \(x_t\) represents an original image \(x_0\) with noise at random time step \(t\). For more detailed information, refer to the method in \sref{sec:dcube} and the analysis in \fref{fig:layer_metric}. The dashed line indicates the optional use of sub features.}
    \label{fig:overall_architecture}
\end{figure*}

\section{Related work}
In this section, we introduce research related to our study, focusing on the adaptation of advanced vision techniques to the medical domain, particularly through the use of diffusion features. We discuss the challenges of integrating traditional vision models into medical imaging and highlight how diffusion can enhance diagnostic accuracy.

\subsection{Medical Domain Adaptation}
As computer vision technology advances, its application in the medical domain has expanded significantly. However, applying well-known general models like ResNet \cite{he2016deep}, ViT \cite{dosovitskiy2021an}, and PiT \cite{heo2021rethinking} to medical domains presents substantial challenges, particularly in distinguishing sensitive features such as fine blood vessels or ambiguous tumors. Extensive research has been conducted on performing segmentation, detection, and classification tasks in the medical image domain \cite{Ashurov, Ling}. Yet, diagnostic models often depend on precisely segmented masks to identify cancer or tumor locations—a process that is both costly and labor-intensive \cite{chen2021applications,braman2017intratumoral,chen2023pancreatic,ren2022convolutional,zhu2019deep, zhang2023robust}. Consequently, there has been a surge in research focused on effectively learning from the abundant unlabeled data in medical domains \cite{systematic}. With the recent emergence of generative models, numerous attempts have been made to apply diffusion algorithms to the medical domain for tasks such as tumor segmentation and prediction \cite{survey, yang, diff_med1,diff_med3, diff_med5}. Efforts have also been made to address data imbalance by generating synthetic data and incorporating them into models \cite{Hardy, azizi2023synthetic}. However, there are situations where it is necessary to use diffusion models in an ensemble for training or to enhance performance further \cite{diff_med2}. Due to the nature of medical data, where small samples often contain significant variability and heterogeneity, training performance can be inconsistently robust, heavily influenced by these data distributions \cite{Dushatskiy}.

\subsection{Exploiting Diffusion Features}
Since the advent of Denoising Diffusion Probabilistic Models (DDPM) \cite{ho2020denoising}, the field of generative diffusion models has witnessed significant progress \cite{chai2023stablevideo,kim2019u}. This advancement has spurred research on applying diffusion algorithms to downstream vision tasks. For instance, DFormer \cite{Wang} effectively employs diffusion models for image segmentation, and recent developments include zero-shot classification, leveraging the generation of random noise from diffusion models \cite{zeroshot}. Additionally, recent studies have analyzed the internal representations of diffusion model features, highlighting their utility in tasks like medical segmentation \cite{Medsegdiff, Chen}. Drawing parallels with DINO, known for its high-quality features used in segmentation and classification, extensive research is underway. This research explores the semantic content of features generated by diffusion algorithms in vision tasks and the use of features from high-performing generative diffusion models \cite{hyperfe,Freedom}. However, some studies focus on aggregating features from a hand-selected subset of layers and specific timesteps $t$ for enhanced semantic understanding \cite{baranchuk2022labelefficient}.

In contrast to these approaches, the DiffMIC algorithm \cite{yang} uses a diffusion model to generate classification logits instead of images. It extracts global and local features using a pretrained ResNet and aggregates them using a top-k method to produce logit vectors from three distinct sets: global features, local features, and a fusion of both. These logits are used as conditioning inputs for the diffusion model, which predicts them during training using mean squared error (MSE) and maximum mean discrepancy (MMD) loss functions. During testing, the model generates logits through sampling, which are then used for classification.

Our model differs significantly from DiffMIC. While DiffMIC focuses on generating classification logits, our model generates images to ensure the semantic richness of features. We extract global features from ResNet, aggregate them via summation to obtain logit vectors, and multiply these logits with single-channel diffusion features. This approach allows us to leverage the diffusion process more integrally, bypassing the need for direct logit-based classification. Furthermore, our study introduces a robust feature selection method for diffusion models, ensuring the selection of semantically rich and meaningful features. This strategy enhances the overall effectiveness of diffusion models in various downstream tasks, especially in complex domains such as medical imaging. By focusing on effective feature selection and integration, our approach aims to improve the performance and applicability of generative diffusion models.

\section{D-Cube}
\label{sec:dcube}
Our training framework is structured around a dual-stage process, as illustrated in \fref{fig:overall_architecture}. 
\subsection{Preliminary: Diffusion Models}
In this study, we employ the DDPM framework, renowned for its exceptional ability to comprehend training data, to extract significant features from medical images. We explain the fundamental operations of the DDPM here.

A diffusion model is probabilistic generative model designed to learn data distributions. The DDPM refines noisy data through a two-step process: forward and reverse. According to Markov chain theory, each step depends solely on the previous step's state. In the forward process, noise is incrementally added at each timestep \( t \) to the original data \( x_0 \), transforming it into \( x_t \), until reaching a predefined large number of steps \( T \), ultimately converting \( x_T \) into Gaussian noise. Conversely, the reverse process aims to reconstruct \( x_0 \) from \( x_t \), wherein the diffusion model learns to denoise the data added during the forward process.

The forward process can be represented as a Markov chain, as shown in \eref{eq:diff_forward}:
\begin{equation}
    \label{eq:diff_forward}
    q(\mathbf{x}_t | \mathbf{x}_{t-1}) = \mathcal{N}(\mathbf{x}_t; \sqrt{\alpha_t} \mathbf{x}_{t-1}, (1 - \alpha_t) \mathbf{I}),
\end{equation}
where $\alpha_t$ is a fixed variance schedule. Our objective is to employ a class-conditional diffusion model to generate data more representative of specific classes.
In reverse process, \( x_t \) is accompanied by a class condition, \( y \).
With learned functions \( \mu_{\theta} \) and \( \Sigma_{\theta} \) representing mean and covariance, the reverse process is expressed as shown in \eref{eq:diff_reverse}:
\begin{equation}
\label{eq:diff_reverse}
p(\mathbf{x}_{t-1} | \mathbf{x}_t, y) = \mathcal{N}(\mathbf{x}_{t-1}; \mu_{\theta}(\mathbf{x}_t, t, y), \Sigma_{\theta}(\mathbf{x}_t, t, y))
\end{equation}

The diffusion loss in our study is designed following the principles outlined for class-conditional diffusion models. Specifically, it is calculated by optimizing the evidence lower bound (ELBO). For a given timestep $t$, input $x$, and class condition $y$, the loss is defined as the MSE between the model's output noise $\epsilon_{\theta}$ given $y$ and the random Gaussian noise $\epsilon$ used to generate $x_t$ at timestep $t$ as expressed in the following \eref{eq:epsilon}:
\begin{equation}
\label{eq:epsilon}
L_{\text{Diff}}(x_t, y) = \mathbb{E}_{t,\epsilon} \left[ \lVert \epsilon - \epsilon_{\theta}(x_t, y) \rVert_2 \right]
\end{equation}
This formulation ensures that the diffusion model learns to minimize the difference between the predicted and true noise, thereby effectively capturing the underlying data distribution.

\subsection{Dual-Stage Training}
\textbf{Step 1} involves training the diffusion model to effectively address challenges such as class indistinguishability and data imbalance prevalent in medical datasets, optimizing the model for robust feature representation. In \textbf{step 2}, we freeze the pretrained diffusion model and then focus on selecting the most effective features from it for classification. Additionally, step 2 introduces advanced techniques like cycle loss and consistency regularization, aimed at further refining the classifier’s performance and ensuring its stability.

\subsubsection{\textbf{Step 1, Training a Diffusion Model}}
\paragraph{\textbf{Objective}} 
Medical datasets often face challenges like class indistinguishability and data imbalance. In such environments, the diffusion model may struggle to learn tail distributions effectively, leading to suboptimal feature representations \cite{qin2023class}. Moreover, when the data volume is limited, the distribution learning process of the model can be time-consuming.
We introduce a contrastive loss function for the diffusion model, making it possible to utilize
the label-aligned representations of the diffusion model to enhance feature representation for the downstream tasks, thereby promoting the embedding of critical semantic information. Our contrastive loss in \eref{eq:Contrastive loss} is designed to minimize feature distances within the same class and maximize them for different classes, thereby fostering robust feature representations for classification.
\begin{equation}
\begin{aligned}
     \label{eq:Contrastive loss}
     L_{\text{Cont}}(x^{1}_t, x^{2}_t)
 = \sum_{}^{N} \bigg[ (1 - \text{target}) \cdot \left\| \text{f}_{\text{mid}}(x^{1}_t) - \text{f}_{\text{mid}}(x^{2}_t) \right\|_2 \\ 
     + \text{target} \cdot \left( \max(\text{margin} - \left\| \text{f}_{\text{mid}}(x^{1}_t) - \text{f}_{\text{mid}}(x^{2}_t) \right\|_2, 0) \right) \bigg]
\end{aligned}
\end{equation}
In \eref{eq:Contrastive loss}, $N$ denotes the batch size, and $t$ indicates a specific timestep, where $x^1_t$ and $x^2_t$ are input data with added noise at timestep $t$. $\text{f}_{\text{mid}}$ represents the middle layer of the diffusion model. $\text{f}_{\text{mid}}(x^1_t)$ is the output from the middle layer when input $x^1_t$ is provided. $\text{f}_{\text{mid}}(x^2_t)$ is defined similarly. When the labels of the two input data are the same, the \text{target} value is 0, and it is 1 otherwise. Hence, $L_{\text{Cont}}$ minimizes the L2 distance between the two feature embeddings when the labels are identical and ensures that the L2 distance exceeds a specific margin when they are different.

The final loss \( L_{\text{Gen}} \) for training the diffusion model is formed as \eref{eq:LGen}. \( x^{1}_t \) and \( x^{2}_t \) represent noisy images at the same timestep \(t\) with labels \( y^{1} \) and \( y^{2} \). The labels \( y^{1} \)  and \( y^{2} \) can be identical or different. \( {\epsilon}^{1} \),\( {\epsilon}^{2} \) represent the predicted noise for \( x^{1}_t \) and \( x^{2}_t \), respectively. The differences between \( {\epsilon}^{1} \) and \( {\epsilon}^{2}\) and the ground truth \( \epsilon \) are calculated as $L_{\text{Diff}}(x^{1}_t)$ and $L_{\text{Diff}}(x^{2}_t)$. This leads to a lower Fr\'{e}chet Inception Distance (FID)~\cite{parmar2021cleanfid}, indicating reduced discrepancy between real and generated data across all classes.
\begin{equation}
    \label{eq:LGen}
    L_{\text{Gen}} = L_{\text{Cont}} + L_{\text{Diff}}(x^{1}_t) + L_{\text{Diff}}(x^{2}_t)
\end{equation}
\subsubsection{\textbf{Step 2, Training a Classifier Model}}
We freeze all layers of the pretrained diffusion model and extract features from specific layers for classification. To enhance this process, we propose \textit{a feature selection method} based on a gaussianity metric to identify the most discriminative features for classification. Additionally, to further improve performance, we introduce a cycle loss and incorporate a consistency regularization loss to stabilize the training process.

\paragraph{\textbf{Feature Selection with a Gaussianity Metric}}
Diffusion models are designed to denoise $x_t$ by accurately predicting the true noise present in $x_t$. This leads to a situation where certain layers in the model produce feature maps that closely resemble a Gaussian distribution, while other layers capture the semantic features of $x_t$. To quantify this phenomenon, we employ the Kolmogorov-Smirnov (KS) test as a Gaussianity metric, which measures the degree to which a feature map approximates a Gaussian distribution. Consequently, if the feature map of a particular layer is found to be close to a Gaussian distribution, it can be inferred that this layer is primarily involved in noise prediction. Conversely, if the feature map deviates from a Gaussian distribution, it indicates that the layer is preserving the semantic features of the image. Based on this understanding, we propose \textit{a feature selection method} to appropriately choose features from the diffusion model for classification, presenting criteria for their use.

Specifically, we employ the KS test to identify and select feature maps that deviate from a Gaussian distribution, as these are more likely to preserve semantic information useful for classification. The KS test compares the cumulative distribution function (CDF) of the feature map with that of a Gaussian distribution. As stated in \eref{eq:ECDF}, The test statistic $D$ is defined as the maximum distance between these two CDFs:

\begin{equation}
\begin{aligned}
    \label{eq:ECDF}
    D = \sup_{x_0} | F_{n,i}(x_0) - F(x_0) |
\end{aligned}
\end{equation}
where $F_{n,i}(x_0)$ is the empirical CDF of the feature map from layer $i$ based on the batch size $n$, and $F(x_0)$ is the CDF of the Gaussian distribution. Here, $x_0$ represents the feature map from a diffusion layer. The p-value obtained from the KS test indicates the likelihood that the feature map follows a Gaussian distribution. If the p-value is greater than 0.05, we consider the feature map to be Gaussian. This detailed analysis enables us to discern which layers are more likely to preserve semantic features versus those primarily engaged in noise prediction, thereby guiding the feature selection process for optimal classification performance.

\paragraph{\textbf{Objective}} 
We propose a cycle loss to improve performance of the classification model.
D-Cube predicts $\Tilde{y}$, and its ground truth $y$ passes through a diffusion model with the same input, generating each output noise.
To minimize the difference in above each noise, the learning process should be directed towards making $\Tilde{y}$ and $y$ same. This contributes to the D-Cube module producing a more accurate $\Tilde{y}$. We utilize cycle loss only to update the weights of the D-Cube so that the pretrained diffusion model from step 1 is kept in a frozen state.
\begin{equation}
\label{eq:Lcycle}
    L_{\text{Cycle}}(x_t, y, \Tilde{y}) = \mathbb{E}\left[ \left\| \epsilon_{\theta}(x_t, \Tilde{y}) - \epsilon_{\theta}(x_t, y) \right\|_2 \right]
\end{equation}
\eref{eq:Lcycle} represents our proposed cycle loss, where \( \theta \) denotes pretrained diffusion model. It quantifies L2 distance of the $\epsilon$ predicted by the model (\( \theta \)) when input $x_t$ and predicted class \( \Tilde{y} \) from step 2 of the training process are fed along with the ground truth class \( y \) at a specific timestep \( t \).

By incorporating consistency regularization loss, we aimed to stabilize the training process by ensuring that the classification output remains consistent even when the input images undergo transformations. Specifically, we trained the D-Cube using both the original images and their horizontally flipped versions, matching the logits produced by these variations. 
The consistency regularization loss, \( L_{\text{CR}} \), is calculated as shown in \eref{eq:LConsistency}. This equation represents the MSE loss between the logits:
\begin{equation}
\label{eq:LConsistency}
    L_{\text{CR}}  = \frac{1}{N} \sum_{i=1}^{N} \left\| g(x_i) - g(T(x_i)) \right\|_2^2
\end{equation}
Here, \( N \) is the batch size, \( x_i \) represents the original image, and \( T(x_i) \) denotes the horizontally flipped image. The function \( g(x) \) is the logit output of the model for the input \( x \).
Cross-entropy loss \( L_{\text{CE}} \) is calculated between predicted class and ground truth classes \( \Tilde{y} \) and \( y \) as shown in \eref{eq:ce}. We scale these losses \( L_{\text{CR}} \) and \( L_{\text{CE}} \) with scaling parameter \( \lambda_1 \) and \( \lambda_2 \). Ultimately, \( L_{CLS} \) becomes the final classification loss as shown in \eref{eq:LDiag}: 
\begin{equation}
\label{eq:ce}
    L_{CE}(y, \tilde{y}) = -\mathbb{E}\left[ y \log(\tilde{y}) + (1 - y) \log(1 - \tilde{y}) \right]
\end{equation}
\begin{equation}
\label{eq:LDiag}
L_{\text{CLS}} = L_{\text{CE}} + \lambda_1L_{\text{Cycle}} + \lambda_2L_{\text{CR}}
\end{equation}

\paragraph{\textbf{Additional Technique}} 
For classification tasks, we leverage features extracted from a pretrained diffusion model. Features selected based on the Gaussianity metric are fed forward into our classifier, D-Cube. To further enhance performance efficiently, we utilize a ResNet pretrained on ImageNet \cite{Deng} to extract sub features for performance improvement, as pretrained ResNet models have been proven to be effective in various downstream tasks \cite{BIT, Yamada}. We obtain a sub feature \(f_{\text{sub}} \in \mathbb{R}^{C \times H \times W}\), corresponding to the arbitrary channel. Features \(f_{\text{sub}}\) and \(f_{\text{main}}\) are concatenated and subjected to global sum pooling to derive sub scores for each channel. Multiplying the single-channel diffusion feature by the ResNet sub features, which we arbitrarily set to three channels, effectively expands the single-channel feature into three channels. This process reflects the important elements identified by the ResNet, thereby achieving the effect of channel expansion. ResNet, along with D-Cube, is fine-tuned to our data, producing \(f_{\text{sub}}\) that is tailored to our specific dataset. These scores are then used to weight the concatenated features. D-Cube incorporates convolutional layers equipped with multiple kernel sizes of 
3 × 3 and 5 × 5, repeated three times. We avoid pooling to preserve spatial information, focusing on a gradual reduction in the channel dimensions of the output. The final prediction is made through two fully connected layers.

\section{Experiments}
\subsection{Datasets}
To facilitate validation across different races and modalities, we used the Duke MRI dataset, representing Western populations, and the National Information Society Agency (NIA) CT dataset for an Asian populations. This strategy aimed to broaden the diversity and global applicability of our research.

\subsubsection{\textbf{NIA Pancreas Cancer CT}}
We utilized a rigorously labeled dataset from the NIA's Medical Big Data Construction Project, comprising CT images of pancreatic cancer and normal tissues from eight South Korean hospitals. Additional data were sourced from the Medical Segmentation Decathlon and the Cancer Imaging Archive. The dataset, encompassing records from 6,063 pancreatic cancer patients, was divided into training and testing sets at a 7:3 ratio. Initially containing 11,775 benign, 11,486 malignant, and 41,033 normal images, we refined the dataset by removing non-pancreatic slices and cropping images to 256 × 256 pixels, centering on the pancreas. After processing, the dataset consisted of 5,513 benign, 6,744 malignant, and 24,626 normal images.

\subsubsection{\textbf{Duke Breast Cancer MRI} \cite{Saha}} 
This dataset comprises scans from 922 patients, with a focus on classifying lymphadenopathy and suspicious lymph nodes in MRI scans from 127 patients. These patients were categorized into training (100 patients) and testing (27 patients) groups based on their IDs. The training dataset included 2,477 images, consisting of 1,071 images classified as lymphadenopathy and 1,406 images classified as suspicious lymph nodes. The testing dataset consisted of 707 images, with 272 images in the lymphadenopathy class and 434 images in the suspicious lymph node class.

\subsubsection{\textbf{COVID-19 Radiography Dataset} \cite{Rahman, Chowdhury}}  We used a chest X-ray dataset that was developed by a team of researchers from Qatar University, Doha, Qatar, and the University of Dhaka, Bangladesh, in collaboration with their counterparts from Pakistan and Malaysia and medical doctors. This comprehensive dataset includes images of COVID-19 positive cases, normal cases, and cases of viral pneumonia. Specifically, the dataset comprises 3,616 COVID-19 positive cases, 10,192 normal cases, 6,012 lung opacity cases (non-COVID lung infections), and 1,345 viral pneumonia cases. The dataset is split into training and testing sets in an 8:2 ratio.

\begin{table*}[t]
\centering
\setlength{\tabcolsep}{6pt}
\caption{\textbf{Quantitative analysis} on pancreas cancer CT, breast cancer MRI, and COVID chest X-ray datasets. The highest and second-highest scores are highlighted in bold and underlined.}
\label{tab:comparison}
\begin{tabular}{l cccc cccc cccc}
\toprule
\multirow{2}{*}{\textbf{Model}} 
& \multicolumn{4}{c}{\textbf{Pancreas Cancer (CT)}} 
& \multicolumn{4}{c}{\textbf{Breast cancer (MRI)}} 
& \multicolumn{4}{c}{\textbf{COVID Chest (X-ray)}} \\
\cmidrule(lr){2-5} \cmidrule(lr){6-9} \cmidrule(lr){10-13}
& \textbf{Acc} & \textbf{Precis} & \textbf{Recall} & \textbf{F1} & 
\textbf{Acc} & \textbf{Precis} & \textbf{Recall} & \textbf{F1} &
\textbf{Acc} & \textbf{Precis} & \textbf{Recall} & \textbf{F1} \\
\midrule
\textbf{CNN-Based} & & & & & & & & & & & & \\
ResNet-101 \cite{he2016deep} & 85.23  & 77.47 & 74.51 & 75.72 & 64.50 & 64.32 & 55.93 & 52.44 & 90.83 &93.20 & 92.37 & 92.71 \\
ResNeXt-101 \cite{xie2017aggregated} & 87.55 & 74.58 & 67.46 & 69.37 & 61.53 & 56.67 & 38.53 & 28.28 & 89.20 & 90.89 & 89.12 & 89.97 \\
ResNeSt \cite{zhang2022resnest} & 87.20 & 81.27 & 76.65 & 78.36 & 66.76 & 43.35 & 40.74 & 40.59 & 92.01 & 93.90 & 92.60 & 93.23 \\
ShuffleNet V2 \cite{ma2018shufflenet} & 89.25 & 84.52 & 81.11 & 82.47  & 68.32 & 66.68 & 63.59 & 63.83 & 90.34 & 92.19 & 91.40 & 91.76 \\
\midrule
\textbf{Transformer-Based} & & & & & & & & & & & &  \\
T2T-ViT \cite{yuan2021tokens} & 82.14 & 74.58 & 67.46 & 69.37 & 62.18 & 82.58 & 57.51 & 52.51 & 87.07 & 90.07 & 85.63 & 87.64 \\
ViT-B/16 \cite{dosovitskiy2021an} & 78.55 & 68.15 & 60.68 & 63.31 & 66.90 & 82.48 & 57.14 & 51.88 & 85.21 & 89.14 & 84.33 & 86.50 \\
CVT \cite{wu2021cvt} & 78.32 & 65.82 & 59.09 & 60.34 & 68.60 & 79.10 & 59.68 & 56.49 & 85.85 & 87.14 & 84.67 & 85.85 \\
MiT-b0 (SegFormer) \cite{xie2021segformer} & 88.77  & 83.88 & 79.87  & 81.63  & 66.62 & 77.18 & 57.12 & 52.34 & 91.49 & 93.62 & 90.07 & 91.71 \\
PiT-B \cite{heo2021rethinking} & 88.24 & 83.21  & 79.18 & 80.71  & 65.91 & 67.98 & 57.42 & 54.33 & 92.60 & 94.29 & 92.72 & 93.48 \\
PVT V2-b0 \cite{wang2022pvt} & 89.59 & 85.21  & 81.33 & 83.09 & 65.49 & 50.47 & 37.10 & 33.22 & 92.41 & 94.04 & 92.32 & 93.15 \\
PVT-Tiny \cite{wang2021pyramid} & 89.52 & 85.96 & 80.31 & 82.47 & 67.47 & 73.16 & 58.83 & 55.82 & 92.93 & 94.84 & 92.84 & 93.80 \\
CPT \cite{viriyasaranon2023unsupervised} & \underline{92.78} & 90.70 & 86.48 & 88.22 & 74.54 & \underline{73.37} &  \underline{71.79} &  \underline{72.30} & 96.08 & 97.18 & \underline{96.50} & 96.83 \\
\midrule
\textbf{Diffusion-Based}  & & & & & & & & \\
DiffMIC \cite{yang}  &  92.68 &  \underline{90.74} &  \underline{86.90} &  \underline{88.53} &  \underline{75.01} & 72.85 & 68.09 & 68.70 & \textbf{96.40} & \textbf{97.72} & \textbf{96.68} & \textbf{97.20} \\
 D-Cube (ours) &  \textbf{93.61} &  \textbf{92.05} &  \textbf{88.05} &  \textbf{89.69} &  \textbf{77.98} &  \textbf{77.87} &  \textbf{74.64} &  \textbf{75.52}  &  \underline{96.28} &   \underline{97.49} &  96.28 &  \underline{96.87} \\
\bottomrule
\end{tabular}
\end{table*}

\subsection{Implementation Details}
We developed our framework using PyTorch and executed it on an NVIDIA RTX A6000 GPU.
\subsubsection{\textbf{Details Regarding Step 1}}
We set the contrastive loss margin to 0.1 and did not perform any additional data augmentation during the first step of training. For the pancreas, breast, and COVID datasets, we set the learning rate at \(2 \times 10^{-4}\) and trained the models for approximately 280, 950, and 730 epochs, respectively, until the FID scores stabilized. We employed the AdamW optimizer. Considering both training speed and memory efficiency, we adopted U-ViT \cite{bao2023all} as our backbone model.

\subsubsection{\textbf{Details Regarding Step 2}}
In step 2, the model uses the original image \( x_0 \) without any noise at a random time step \( t \) as input. During the classifier module training, we applied random horizontal flip augmentation with a 50\% probability. For the pancreas, breast, and COVID datasets, we set the learning rate at \(2 \times 10^{-4}\). Due to the relatively small size of the breast dataset, we stabilized training by setting a weight decay of 0.1. We used the AdamW optimizer and enhanced training stability by applying the Exponential Moving Average (EMA) with an \(\alpha\) value of 0.999. The pancreas cancer CT and COVID chest X-ray datasets achieved consistent performance around 70 epochs, while the breast cancer MRI dataset continued experiments up to 50 epochs. We scaled the cycle loss by applying weights of 10 for the pancreas and 100 for the breast. When applying consistency regularization loss, we empirically set the \(\lambda\) value to 0.1, achieving the highest performance across all datasets.

\subsection{\textbf{Benchmarking}}
We performed experiments to compare our D-Cube model with existing CNN-based, Transformer-based, and diffusion-based models. In the diffusion-based medical classification models, due to the limited number of studies and the lack of publicly available code, reproducing the results was challenging. Therefore, we only compared DiffMIC with open source. 

In the medical domain, metrics like recall and F1 score are essential. High recall ensures that diseases are identified with minimal misses, while the F1 score provides a balanced measure of a model's accuracy, especially important in medical settings to avoid both false positives and false negatives.

\indent As shown in \tref{tab:comparison}, D-Cube surpasses previous models in almost metrics across datasets. This achievement is attributed to the features D-Cube learns, effectively capturing the training data distribution and embedding essential semantic information to boost classification performance. Furthermore, the experimental results illustrate that D-Cube outperforms CPT, a transformer-based architecture model that utilized self-supervised pretraining on medical images domain, achieving the best accuracy among transformer-based models. The DiffMIC, which employs a diffusion algorithm that effectively leverages the training data, demonstrated performance similar to ours; however, D-Cube achieves accuracies of 93.61\% and 77.98\% for pancreas cancer CT and breast cancer MRI datasets respectively, compared to DiffMIC with accuracies of 92.68\% and 75.01\% on the same datasets. Notably, D-Cube outperforms DiffMIC by approximately 6\% on the smaller Breast dataset. Therefore, our D-Cube has demonstrated robust performance even in situations with limited data, regardless of the domain. \\
\indent Our model has the advantage of generating images, which were used to address class imbalance issues in training lower-performing CNN and transformer models, as demonstrated in \tref{tab:comparison}. For the pancreas data, we added 30,000 synthetic images, adjusting the dataset's ratio from 1:1:4 (benign, malignant, and normal) to a more balanced 2:2:3. As a result, as shown in \tref{tab:fake_data}, we observed improvements in all evaluation metrics with notable increases of approximately 2-4\% in recall and F1 score, demonstrating our approach's efficacy.

\begin{table}[t]
\centering
\caption{\textbf{Augmentation with synthetic data} of D-Cube on pancreas cancer CT. Syn indicates generated data from diffusion model of step 1.}
\label{tab:fake_data}
\begin{tabular}{l cccc}
\toprule
\multirow{1}{*}{\textbf{Model}}
& \textbf{Acc} & \textbf{Precis} & \textbf{Recall} & \textbf{F1} \\
\midrule
ResNeXt-101 & 87.55 & 81.87 & 77.94 & 78.92 \\
+ Syn & 88.53\textsuperscript{+0.98} & 83.18\textsuperscript{+1.31} & 79.86\textsuperscript{+1.92} & 80.99\textsuperscript{+2.07} \\
\midrule
ViT-B/16 & 78.55 & 68.15 & 60.68 & 63.31 \\
+ Syn & 80.59\textsuperscript{+2.04} & 70.83\textsuperscript{+2.68} & 64.90\textsuperscript{+4.22} & 67.14\textsuperscript{+3.83} \\
\bottomrule
\end{tabular}
\end{table}

\begin{table}[t]
\centering
\caption{\textbf{Ablation analysis} of D-Cube on pancreas cancer CT. Fs indicates using the proposed Feature Selection, and \(f_{\text{sub}}\) indicates the use of sub features extracted from the ResNet.}
\label{tab:ablation_loss}
\begin{tabular}{l cccc}
\toprule
\multirow{1}{*}{\textbf{Model}}
& \textbf{Acc} & \textbf{Precis} & \textbf{Recall} & \textbf{F1} \\
\midrule
Baseline & 85.30 & 77.82 & 74.81 & 75.82 \\
+ $\text{L}_\text{Gen}$ & 85.97\textsuperscript{+0.67} & 78.95\textsuperscript{+1.13} & 75.91\textsuperscript{+1.1} & 77.10\textsuperscript{+1.28} \\
+ Fs & 87.86\textsuperscript{+1.89} & 82.13\textsuperscript{+3.18} & 78.64\textsuperscript{+2.73} & 79.95\textsuperscript{+2.85} \\
+ $\text{L}_\text{Cls}$ & 88.26\textsuperscript{+0.4} & 82.77\textsuperscript{+0.64} & 79.20\textsuperscript{+0.56} & 80.61\textsuperscript{+0.66} \\
 + \(f_{\text{sub}}\) &  93.61\textsuperscript{+5.35} &  92.05\textsuperscript{+9.28} &  88.05\textsuperscript{+8.85} &  89.69\textsuperscript{+9.08} \\
\bottomrule
\end{tabular}
\end{table}

\begin{figure}[t]
    \centerline{\includegraphics[width=0.7\columnwidth]{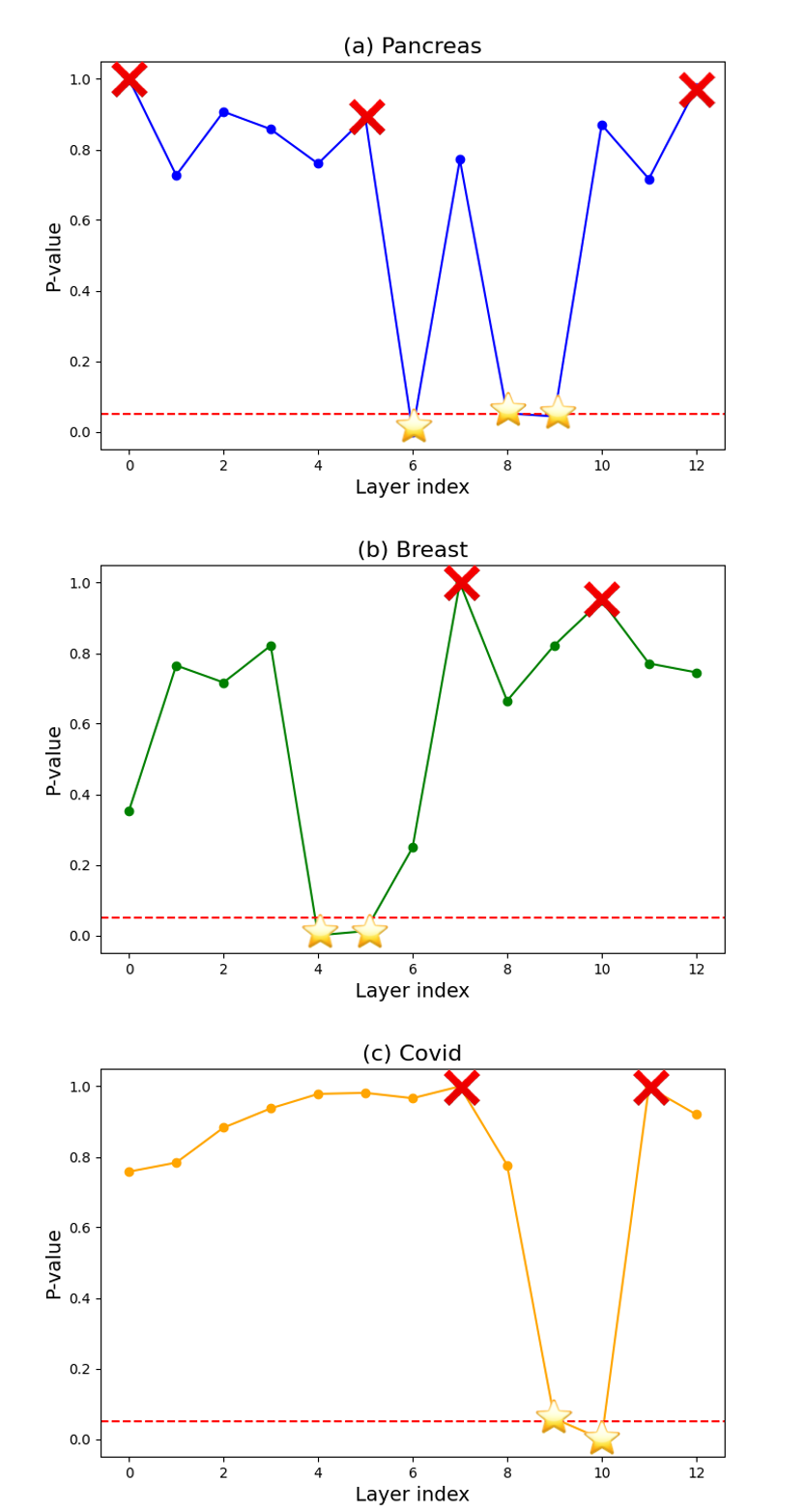}}
    \caption{
    \textbf{Gaussianity Test of Each Datasets} This graph presents the p-value analysis for pancreas cacner CT, breast cancer MRI, and COVID chest X-ray.  Yellow star denote the best combination used, while red cross marks represent the worst combination. The p-value is calculated based on the average of features with a batch size of 256 when t=100.
}
    \label{fig:layer_metric}
\end{figure}

\begin{table*}[t]
\centering
\caption{\textbf{Comparison of FID Scores} Synthetic images generated by a diffusion model with and without $\text{L}_\text{Gen}$ on pancreas cancer CT, breast cancer MRI, and COVID chest X-ray Datasets.}
\label{tab:FID}
\begin{tabular}{l @{\hspace{1em}} cccc @{\hspace{2em}} cccc @{\hspace{2em}} cccc}
\toprule
\multicolumn{10}{c}{\textbf{FID ↓}} \\
\cmidrule(lr){2-10}
& \multicolumn{3}{c}{\textbf{Pancreas Cancer (CT)}} & \multicolumn{2}{c}{\textbf{Breast Cancer (MRI)}} & \multicolumn{4}{c}{\textbf{COVID Chest (X-ray)}} \\
\cmidrule(lr){2-4} \cmidrule(lr){5-6} \cmidrule(lr){7-10}
& \textbf{Benign} & \textbf{Malignant} & \textbf{Normal} & \textbf{Lymphadenopathy} & \textbf{Suspicious} & \textbf{COVID} & \textbf{Normal} & \textbf{Opactiy} & \textbf{Viral} \\
\midrule
U-ViT (step 1) & 24.8 & 29.3 & 27.3 & 79.7 & 71.1 & 70.03 & 58.53 & 74.60 & 112.70 \\
U-ViT (step 1) w/ $\text{L}_\text{Gen}$ & 22.2 & 24.0 & 21.0 & 66.2 & 66.9 & 66.86 & 59.14 & 74.81 & 101.02 \\
\bottomrule
\end{tabular}
\end{table*}

\begin{table*}[t]
\centering
\setlength{\tabcolsep}{6pt}
\caption{\textbf{Feature Selection Analysis} on pancreas cancer CT, breast cancer MRI, and COVID chest X-ray datasets. We selected features based on \fref{fig:layer_metric}.}
\label{tab:selection}
\begin{tabular}{l cccc cccc cccc}
\toprule
\multirow{2}{*}{} 
& \multicolumn{4}{c}{\textbf{Pancreas Cancer (CT)}} 
& \multicolumn{4}{c}{\textbf{Breast Cancer (MRI)}} 
& \multicolumn{4}{c}{\textbf{COVID Chest (X-ray)}} \\
\cmidrule(lr){2-5} \cmidrule(lr){6-9} \cmidrule(lr){10-13} 
& \textbf{Acc} & \textbf{Precis} & \textbf{Recall} & \textbf{F1} & 
\textbf{Acc} & \textbf{Precis} & \textbf{Recall} & \textbf{F1} &
\textbf{Acc} & \textbf{Precis} & \textbf{Recall} & \textbf{F1} \\
\midrule
Worst Selection & 93.23 & 91.25 &  87.38 & 88.94 & 76.84 & 75.63 & 74.78 & 75.12 & 95.85 & 97.28 & 95.59 & 96.41 \\
 Best Selection &  \textbf{93.61} &  \textbf{91.97} &  \textbf{87.92} &  \textbf{89.52} &  \textbf{77.98} &  \textbf{77.87} &  \textbf{74.64} &  \textbf{75.52}  &  \textbf{96.28} &  \textbf{97.49} &  \textbf{96.28} &  \textbf{96.87} \\
\bottomrule
\end{tabular}
\end{table*}

\subsection{Ablation Study}
\subsubsection{\textbf{Model Ablation}}
Our quantitative results confirm that features from the diffusion model, which deeply understand the training data, contain crucial semantic information for classification tasks. We explored this by conducting an ablation study, assessing the benefits of features derived from the diffusion model and the impact of various loss functions on the D-Cube's performance, as shown in \tref{tab:ablation_loss}. For the baseline, in training step 1, we applied only $L_{\text{diff}}$, and in step 2 within the classifier, we applied only $L_{\text{CE}}$. Moreover, we used the features from the final output layer before applying feature selection.  Additionally, the efficacy of a diffusion model is indicated by lower FID scores. Our $L_{\text{Gen}}$ method led to reduced FID scores, as shown in \tref{tab:FID}, thereby enhancing performance. Enhancements in the diffusion model through $L_{\text{Gen}}$ are linked to improved D-Cube performance. Moreover, our feature selection strategies (Fs) helped identify crucial features for classification, enabling high performance even in scenarios with imbalanced and scarce data. In \fref{fig:layer_metric}, features from layers marked with yellow star—corresponding to p-values less than 0.05—were considered the best selections and applied. This approach resulted in significant performance improvements. As observed in \tref{tab:ablation_loss}, the classification performance using features from the diffusion model, without the aid of sub-features, matches that of CNN and transformer based models. This implies that the diffusion model, although not specifically designed for classification tasks, is capable of learning valuable feature representations that enhance classification effectiveness. The integration of our $L_{\text{Cls}}$ allowed for more refined learning, and the use of \(f_{\text{sub}}\) positively influenced D-Cube’s effectiveness. Through strategic feature selection and diverse loss functions, D-Cube's performance, detailed in \tref{tab:comparison}, matches other models except for DiffMIC. By utilizing sub-features, we outperformed DiffMIC.

\subsubsection{\textbf{Feature Selection}}
We utilized a Gaussianity metric to determine which layer feature maps in diffusion models for pancreas cancer CT, breast cancer MRI, and COVID chest X-ray approximate a Gaussian distribution, as shown in \fref{fig:layer_metric}. The diffusion models are designed to accurately predict true noise present in $x_t$. Therefore, if a feature map of a specific layer closely resembles a Gaussian distribution, it implies that this feature map is predominantly noise with minimal semantic information. Conversely, if a feature map deviates from a Gaussian distribution, it indicates that the layer retains significant semantic information in $x_t$. Based on these findings, layers that contain the most semantic information were marked with yellow stars, while those that did not were marked with red crosses.\\
\indent We concatenated the feature maps of the layers marked with yellow stars, referred to as the \textit{Best Selection}, and those marked with red crosses, referred to as the \textit{Worst Selection}. We then conducted a comparative analysis for each dataset, the results of which are presented in \tref{tab:selection}. Our analysis reveals that the \textit{Best Selection} retains more semantic information, leading to superior performance, thereby validating our feature selection.

\begin{table}[t]
\centering
\caption{\textbf{Ablation Study of D-Cube on Pancreatic Cancer CT}. The results are based on variations in input image noise and random timesteps \({t}\) during the classification process in step 2}
\label{tab:ablaiton_input}
\begin{tabular}{l cccc}
\toprule
\multirow{1}{*}{\textbf{Input}}
& \textbf{Acc} & \textbf{Precis} & \textbf{Recall} & \textbf{F1} \\
\midrule
 \( x_0 \), \( t \) &  93.61 &  92.05  & 88.05 & 89.69 \\
\( x_t \), \( t \)   & 93.40 & 91.81 & 87.77 & 89.33\\
\( x_0 \), \( 0 \)  & 93.10 & 91.41 & 87.03 & 88.79\\
\bottomrule
\end{tabular}
\end{table}

\begin{table}[t]
\centering
\caption{\textbf{Sub features  analysis} of D-Cube on pancreas cancer CT.}
\label{tab:sub_features}
\begin{tabular}{l @{\hspace{2em}} cccc}
\toprule
\multirow{1}{*}{\textbf{Model of Sub Features}}
& \textbf{Acc} & \textbf{Precis} & \textbf{Recall} & \textbf{F1} \\
\midrule
\textbf{CNN-Based} & & & & \\
ResNeXt-101 & 93.61 & 92.05 & 88.05 & 89.6 \\
DenseNet & 93.40 & 91.36 & 87.76 & 89.21 \\
\midrule
\textbf{Transformer-Based} & & & &  \\
ViT & 80.94 & 69.70 & 65.83 & 66.75 \\
RADIO & 78.28 & 61.56 & 65.97 & 62.86\ \\
\bottomrule
\end{tabular}
\end{table}

\subsubsection{\textbf{Input Ablation}}
Given the random sampling nature of diffusion-based methods, we analyzed the robustness of our proposed method according to the input. As demonstrated in \tref{tab:ablaiton_input} through input ablation studies, our model showed remarkably robust performance, unaffected by the random Gaussian noise added to the input. Notably, the best performance was observed using the original images \(x_{\text{0}}\) without added noise and at a randomly generated timestep \({t}\).

\subsubsection{\textbf{Compatibility of Pretrained Models with Diffusion Features}}
In contrast to general natural images, medical images require the capture of fine details due to the often small size of lesions. To achieve this, networks need to function like high-pass filters, amplifying high-frequency components. Moreover, the medical domain typically has a limited amount of data, making inductive biases such as locality and translation invariance advantageous for learning from limited data. Vision Transformers (ViTs) tend to reduce high-frequency components, resulting in a shape bias, whereas CNNs amplify high-frequency components, leading to a texture bias \cite{Park}. The inductive bias inherent in CNNs makes them particularly effective in data-constrained environments such as the medical field. Our experiments in \tref{tab:sub_features} confirm that CNN-based models, such as ResNet and DenseNet, are more adept at capturing useful \(f_{\text{sub}}\); however, when using sub-features from the transformer-based models, like ViT and RADIO series, it was observed that the diffusion features did not synergize well.

\section{Conclusion}
In conclusion, our research convincingly demonstrates the advantages of diffusion models in the medical domain, significantly outperforming traditional foundation models. Diffusion models show exceptional capability in addressing the specific challenges of limited data and class imbalance often encountered in medical imaging tasks. By leveraging \textit{feature selection} via the Gaussianity metric, particularly focusing on non-Gaussian features, we have notably enhanced the performance of D-Cube. 

Our approach utilizes the generative capacity of diffusion models to improve the accuracy of networks by incorporating synthetic data and employing tailored loss functions and feature selection techniques. This strategy has resulted in state-of-the-art performance across various medical imaging modalities. Additionally, our findings reveal that diffusion features, when combined with sub-features from CNN architectures, effectively capture fine details essential for accurately detecting subtle anatomical structures in medical images. 

For future work, we aim to extend our insights and methodologies to general datasets beyond medical diagnostics, exploring the potential of diffusion models in other fields characterized by data scarcity and complexity. By continuing to refine feature selection criteria and integrating advanced deep learning techniques, we anticipate further advancements in the performance and applicability of diffusion models in diverse domains.

\section{Acknowledgement}
This work was supported by Institute of Information \& communications Technology Planning \& Evaluati
on (IITP) grant funded by the Korea government(MSIT). (No. RS-2022-00155966, Artificial Intelligence Convergence Innovation Human Resources Development (Ewha Womans University))

\newpage
\appendix
\section{Appendix.}

\subsection{Implementation details}
\tref{tab:implementation_details} displays the hyperparameters for benchmarking models applied to three datasets: pancreas cancer CT, breast cancer MRI, and covid chest X-ray. The same hyperparameters were used across all datasets, and the number of epochs was determined based on the convergence of accuracy.

\begin{table}[t]
\centering
\caption{\textbf{The network configuration for benchmarking}}
\label{tab:implementation_details}
\begin{tabular}{@{}lcccc@{}}
\toprule
Method    & Learning Rate & Weight Decay & Epochs (\#) & Optimizer \\
\midrule
ResNet-101 & 1e-3         & 1e-4       & 50       & SGD       \\
ResNeXt-101 & 1e-3        & 1e-4       & 50       & SGD       \\
ShuffleNet V2 & 1e-3      & 5e-5      & 50       & SGD       \\
ViT-Base/16 & 6e-6         & 3e-1         & 200      & AdamW     \\
MiT-b0 (SegFormer) & 8e-6  & 1e-2        & 100      & AdamW     \\
CVT        & 6.5e-5       & 1e-2          & 100      & AdamW     \\
PVT v2 - b0 & 1e-5         & 1e-2         & 50       & AdamW     \\
PVT - tiny & 8e-6         & 1e-2          & 50       & AdamW     \\
CPT (ours) & 5e-5         & 1e-2          & 50       & AdamW     \\ \bottomrule
\end{tabular}
\end{table}


\begin{thebibliography}{00}

\bibitem{unet} O. Ronneberger, P. Fischer, and T. Brox, ``U-net: Convolutional networks for biomedical image segmentation,'' in Proc. MICCAI, 2015.

\bibitem{chen2020synthetic} L. Chen, X. Liang, C. Shen, S. Jiang, and J. Wang, ``Synthetic CT generation from CBCT images via deep learning,'' Med. Phys., 2020.

\bibitem{ho2020denoising} J. Ho, A. Jain, and P. Abbeel, ``Denoising diffusion probabilistic models,'' NeurIPS, 2020.


\bibitem{chen2021applications} P.-T. Chen, D. Chang, T. Wu, M.-S. Wu, W. Wang, and W.-C. Liao, ``Applications of artificial intelligence in pancreatic and biliary diseases,'' J. Gastroenterol. Hepatol., 2021.

\bibitem{chen2023pancreatic} P.-T. Chen, T. Wu, P. Wang, D. Chang, K.-L. Liu, M.-S. Wu, H. R. Roth, P.-C. Lee, W.-C. Liao, and W. Wang, ``Pancreatic cancer detection on CT scans with deep learning: a nationwide population-based study,'' Radiology, 2023.

\bibitem{viriyasaranon2023unsupervised} T. Viriyasaranon, S. M. Woo, and J.-H. Choi, ``Unsupervised Visual Representation Learning Based on Segmentation of Geometric Pseudo-Shapes for Transformer-Based Medical Tasks,'' IEEE J. Biomed. Health Inform., 2023.

\bibitem{zhang2023robust} Zhang, Jiadong, Zhiming Cui, Zhenwei Shi, Yingjia Jiang, Zhiliang Zhang, Xiaoting Dai, Zhenlu Yang, Yuning Gu, Lei Zhou, Chu Han, et al. "A robust and efficient AI assistant for breast tumor segmentation from DCE-MRI via a spatial

\bibitem{azizi2023synthetic} S. Azizi, S. Kornblith, C. Saharia, M. Norouzi, and D. J. Fleet, ``Synthetic data from diffusion models improves imagenet classification,'' arXiv preprint arXiv:2304.08466, 2023.

\bibitem{bao2023all} F. Bao, S. Nie, K. Xue, Y. Cao, C. Li, H. Su, and J. Zhu, ``All are worth words: A vit backbone for diffusion models,'' in Proc. CVPR, 2023.

\bibitem{dalle} A. Ramesh, M. Pavlov, G. Goh, S. Gray, C. Voss, A. Radford, M. Chen, and I. Sutskever, ``Zero-shot text-to-image generation,'' in Proc. ICML, 2021.

\bibitem{chai2023stablevideo} W. Chai, X. Guo, G. Wang, and Y. Lu, ``Stablevideo: Text-driven consistency-aware diffusion video editing,'' in Proc. ICCV, 2023.

\bibitem{he2016deep} K. He, X. Zhang, S. Ren, and J. Sun, ``Deep residual learning for image recognition,'' in Proc. CVPR, 2016.

\bibitem{xie2017aggregated} S. Xie, R. Girshick, P. Doll{\'a}r, Z. Tu, and K. He, ``Aggregated residual transformations for deep neural networks,'' in Proc. CVPR, 2017.

\bibitem{zhang2022resnest} H. Zhang, C. Wu, Z. Zhang, Y. Zhu, H. Lin, Z. Zhang, Y. Sun, T. He, J. Mueller, R. Manmatha, and others, ``ResNeSt: Split-Attention Networks'' in Proc. CVPR, 2022.

\bibitem{ma2018shufflenet} N. Ma, X. Zhang, H.-T. Zheng, and J. Sun, ``Shufflenet v2: Practical guidelines for efficient cnn architecture design,'' in Proc. ECCV, 2018.

\bibitem{dosovitskiy2021an} A. Dosovitskiy, L. Beyer, A. Kolesnikov, D. Weissenborn, X. Zhai, T. Unterthiner, M. Dehghani, M. Minderer, G. Heigold, S. Gelly, J. Uszkoreit, and N. Houlsby, ``An Image is Worth 16x16 Words: Transformers for Image Recognition at Scale,'' in Proc. ICLR, 2021.

\bibitem{yuan2021tokens} L. Yuan, Y. Chen, T. Wang, W. Yu, Y. Shi, Z.-H. Jiang, F. E. Tay, J. Feng, and S. Yan, ``Tokens-to-token vit: Training vision transformers from scratch on imagenet,'' in Proc. ICCV, 2021.

\bibitem{wu2021cvt} H. Wu, B. Xiao, N. Codella, M. Liu, X. Dai, L. Yuan, and L. Zhang, ``Cvt: Introducing convolutions to vision transformers,'' in Proc. ICCV, 2021.

\bibitem{heo2021rethinking} B. Heo, S. Yun, D. Han, S. Chun, J. Choe, and S. J. Oh, ``Rethinking spatial dimensions of vision transformers,'' in Proc. ICCV, 2021.

\bibitem{wang2022pvt} W. Wang, E. Xie, X. Li, D.-P. Fan, K. Song, D. Liang, T. Lu, P. Luo, and L. Shao, ``Pvt v2: Improved baselines with pyramid vision transformer,'' Computational Visual Media, 2022.

\bibitem{wang2021pyramid} W. Wang, E. Xie, X. Li, D.-P. Fan, K. Song, D. Liang, T. Lu, P. Luo, and L. Shao, ``Pyramid vision transformer: A versatile backbone for dense prediction without convolutions,'' in Proc. ICCV, 2021.

\bibitem{parmar2021cleanfid} G. Parmar, R. Zhang, and J.-Y. Zhu, ``On Aliased Resizing and Surprising Subtleties in GAN Evaluation,'' in Proc. CVPR, 2022.

\bibitem{xie2021segformer} E. Xie, W. Wang, Z. Yu, A. Anandkumar, J. M. Alvarez, and P. Luo, ``SegFormer: Simple and efficient design for semantic segmentation with transformers,'' NeurIPS, 2021.

\bibitem{mri2019overview} A. S. Lundervold and A. Lundervold, ``An overview of deep learning in medical imaging focusing on MRI,'' Zeitschrift für Medizinische Physik, 2019.

\bibitem{kim2019u} J. Kim, M. Kim, H. Kang, and K. H. Lee, ``U-GAT-IT: Unsupervised Generative Attentional Networks with Adaptive Layer-Instance Normalization for Image-to-Image Translation,'' in Proc. ICLR, 2019.

\bibitem{braman2017intratumoral} N. M. Braman, M. Etesami, P. Prasanna, C. Dubchuk, H. Gilmore, P. Tiwari, D. Plecha, and A. Madabhushi, ``Intratumoral and peritumoral radiomics for the pretreatment prediction of pathological complete response to neoadjuvant chemotherapy based on breast DCE-MRI,'' Breast Cancer Res., 2017.

\bibitem{ren2022convolutional} T. Ren, S. Lin, P. Huang, and T. Q. Duong, ``Convolutional neural network of multiparametric MRI accurately detects axillary lymph node metastasis in breast cancer patients with pre neoadjuvant chemotherapy,'' Clin. Breast Cancer, 2022.

\bibitem{zhu2019deep} Z. Zhu, E. Albadawy, A. Saha, J. Zhang, M. R. Harowicz, and M. A. Mazurowski, ``Deep learning for identifying radiogenomic associations in breast cancer,'' Comput. Biol. Med., 2019.

\bibitem{yang} Yang, Y., Fu, H., Aviles-Rivero, A. I., Schönlieb, C. B., \& Zhu, L. (2023, October). Diffmic: Dual-guidance diffusion network for medical image classification. In International Conference on Medical Image Computing and Computer-Assisted Intervention (pp. 95-105). Cham: Springer Nature Switzerland.

\bibitem{Chowdhury} M.E.H. Chowdhury, T. Rahman, A. Khandakar, R. Mazhar, M.A. Kadir, Z.B. Mahbub, K.R. Islam, M.S. Khan, A. Iqbal, N. Al-Emadi, M.B.I. Reaz, M. T. Islam, “Can AI help in screening Viral and COVID-19 pneumonia?” IEEE Access, Vol. 8, 2020, pp. 132665 - 132676.

\bibitem{Rahman} Rahman, T., Khandakar, A., Qiblawey, Y., Tahir, A., Kiranyaz, S., Kashem, S.B.A., Islam, M.T., Maadeed, S.A., Zughaier, S.M., Khan, M.S. and Chowdhury, M.E., 2020. Exploring the Effect of Image Enhancement Techniques on COVID-19 Detection using Chest X-ray Images. arXiv preprint arXiv:2012.02238.


\bibitem{Saha} Saha, A., Harowicz, M.R., Grimm, L.J., Kim, C.E., Ghate, S.V., Walsh, R. and Mazurowski, M.A., 2018. A machine learning approach to radiogenomics of breast cancer: a study of 922 subjects and 529 DCE-MRI features. British journal of cancer, 119(4), pp.508-516.

\bibitem{systematic} Huang, S. C., Pareek, A., Jensen, M., Lungren, M. P., Yeung, S., \& Chaudhari, A. S. (2023). Self-supervised learning for medical image classification: a systematic review and implementation guidelines. NPJ Digital Medicine, 6(1), 74.

\bibitem{diff_med1} Feng, Z., Wen, L., Xiao, J., Xu, Y., Wu, X., Zhou, J., ... \& Wang, Y. (2023). Diffusion-based Radiotherapy Dose Prediction Guided by Inter-slice Aware Structure Encoding. arXiv preprint arXiv:2311.02991.

\bibitem{diff_med2}Zapaishchykova, Anna, et al. "Diffusion deep learning for brain age prediction and longitudinal tracking in children through adulthood." Imaging Neuroscience 2 (2024): 1-14.

\bibitem{diff_med3}Fu, L., Li, X., Cai, X., Wang, Y., Wang, X., Yao, Y., \& Shen, Y. (2023). SP-DiffDose: A Conditional Diffusion Model for Radiation Dose Prediction Based on Multi-Scale Fusion of Anatomical Structures, Guided by SwinTransformer and Projector. arXiv preprint arXiv:2312.06187.


\bibitem{diff_med5} Lv, T., Liu, Y., Miao, K., Li, L., \& Pan, X. (2023, October). Diffusion Kinetic Model for Breast Cancer Segmentation in Incomplete DCE-MRI. MICCAI (pp. 100-109). Cham: Springer Nature Switzerland.

\bibitem{Hardy}Hardy, R., Klepich, J., Mitchell, R., Hall, S., Villareal, J., \& Ilin, C. (2023). Improving nonalcoholic fatty liver disease classification performance with latent diffusion models. Scientific Reports, 13(1), 21619.

\bibitem{Ling}Ling, Y., Wang, Y., Dai, W., Yu, J., Liang, P., \& Kong, D. (2023). MTANet: Multi-Task Attention Network for Automatic Medical Image Segmentation and Classification. IEEE Transactions on Medical Imaging.

\bibitem{Ashurov} Ashurov, A., Chelloug, S. A., Tselykh, A., Muthanna, M. S. A., Muthanna, A., \& Al-Gaashani, M. S. (2023). Improved breast Cancer classification through combining transfer learning and attention mechanism. Life, 13(9), 1945.

\bibitem{Dushatskiy}Dushatskiy, Arkadiy, et al. "Data variation-aware medical image segmentation." Medical Imaging 2022: Image Processing. Vol. 12032. SPIE, 2022.

\bibitem{Medsegdiff}Wu, Junde, et al. "Medsegdiff: Medical image segmentation with diffusion probabilistic model." Medical Imaging with Deep Learning. PMLR, 2024.

\bibitem{Chen}Chen, Shoufa, et al. "Diffusiondet: Diffusion model for object detection." Proceedings of the IEEE/CVF ICCV. 2023.

\bibitem{Wang}Wang, H., Cao, J., Anwer, R. M., Xie, J., Khan, F. S., \& Pang, Y. (2023). Dformer: Diffusion-guided transformer for universal image segmentation. arXiv preprint arXiv:2306.03437.

\bibitem{zeroshot}Li, Alexander C., et al. "Your diffusion model is secretly a zero-shot classifier." Proceedings of the IEEE/CVF International Conference on Computer Vision. 2023.

\bibitem{hyperfe}Luo, Grace, et al. "Diffusion hyperfeatures: Searching through time and space for semantic correspondence." Advances in Neural Information Processing Systems 36 (2024).

\bibitem{baranchuk2022labelefficient} Baranchuk, D., Rubachev, I., Voynov, A., Khrulkov, V.,\& Babenko, A. "Label-Efficient Semantic Segmentation with Diffusion Models." In \textit{ICCV}, 2022.

\bibitem{Freedom}Yu, Jiwen, et al. "Freedom: Training-free energy-guided conditional diffusion model." Proceedings of the IEEE/CVF ICCV. 2023.

\bibitem{Radford}Radford, Alec, et al. "Learning transferable visual models from natural language supervision." ICML. PMLR, 2021.


\bibitem{Dino}Caron, M., Touvron, H., Misra, I., Jégou, H., Mairal, J., Bojanowski, P., \& Joulin, A. (2021). Emerging properties in self-supervised vision transformers. In Proceedings of the IEEE/CVF ICCV.

\bibitem{Deng}Deng, J., Dong, W., Socher, R., Li, L. J., Li, K., \& Fei-Fei, L. (2009, June). Imagenet: A large-scale hierarchical image database. In 2009 IEEE conference on CVPR (pp. 248-255). Ieee.

\bibitem{Park}Park, Namuk, and Songkuk Kim. "How do vision transformers work?." arXiv preprint arXiv:2202.06709 (2022).

\bibitem{Ranzinger}Ranzinger, Mike, Greg Heinrich, Jan Kautz, and Pavlo Molchanov. "AM-RADIO: Agglomerative Visual Foundation Model -- Reduce All Domains Into One." In CVPR, 2024.

\bibitem{BIT}Kolesnikov, A., Beyer, L., Zhai, X., Puigcerver, J., Yung, J., Gelly, S., \& Houlsby, N. (2020). Big transfer (bit): General visual representation learning. In Computer Vision–ECCV 2020: 16th European Conference, 2020.

\bibitem{Yamada}Yamada, Yutaro, and Mayu Otani. "Does robustness on imagenet transfer to downstream tasks?." Proceedings of CVPR. 2022.

\bibitem{Hu2022LoRA}
Hu, Edward J., et al. "LoRA: Low-Rank Adaptation of Large Language Models." ICLR. 2022. 


\bibitem{medclip}Wang, Z., Wu, Z., Agarwal, D., \& Sun, J. (2022). Medclip: Contrastive learning from unpaired medical images and text. arXiv preprint arXiv:2210.10163.

\bibitem{survey}Kazerouni, A., Aghdam, E. K., Heidari, M., Azad, R., Fayyaz, M., Hacihaliloglu, I., \& Merhof, D. (2023). Diffusion models in medical imaging: A comprehensive survey. Medical Image Analysis, 102846.

\bibitem{Jia}Jia, C., Yang, Y., Xia, Y., Chen, Y. T., Parekh, Z., Pham, H., ... \& Duerig, T. (2021, July). Scaling up visual and vision-language representation learning with noisy text supervision. ICML. PMLR.

\bibitem{openai2024sora} OpenAI. ``Sora: Creating video from text.'' Accessed 2024. \url{https://openai.com/sora}

\bibitem{Litjens}Litjens, G., Kooi, T., Bejnordi, B. E., Setio, A. A. A., Ciompi, F., Ghafoorian, M., ... \& Sánchez, C. I. (2017). A survey on deep learning in medical image analysis. Medical image analysis.

\bibitem{Shin}Shin, H. C., Roth, H. R., Gao, M., Lu, L., Xu, Z., Nogues, I., ... \& Summers, R. M. (2016). Deep convolutional neural networks for computer-aided detection: CNN architectures, dataset characteristics and transfer learning. IEEE transactions on medical imaging, 35(5), 1285-1298.

\bibitem{Wongvorachan}Wongvorachan, T., He, S., \& Bulut, O. (2023). A comparison of undersampling, oversampling, and SMOTE methods for dealing with imbalanced classification in educational data mining. Information, 14(1), 54.

\bibitem{qin2023class}Qin, Y., Zheng, H., Yao, J., Zhou, M., \& Zhang, Y. (2023). Class-balancing diffusion models. CVPR.

\end{thebibliography}
\end{document}